\documentclass[letterpaper, 10 pt, journal, twoside]{IEEEtran} 
\usepackage{times}
\usepackage[pdftex]{graphicx}
\usepackage{subfigure}
\usepackage{amsmath,amssymb,amsopn,amstext,amsfonts}
\usepackage{cancel}
\usepackage[space]{cite}
\usepackage{pdfsync}
\usepackage{balance}
\usepackage{color}
\usepackage{mathtools}
\usepackage{bm}

\usepackage{diagbox}
\usepackage{float}
\usepackage{epstopdf}
\usepackage{pifont}
\usepackage{fixltx2e}
\usepackage{amsmath}
\usepackage{multirow}
\usepackage{url}
\usepackage{verbatim}
\usepackage[linkcolor=black,citecolor=black,urlcolor=black,colorlinks=true]{hyperref}
\usepackage{soul,xcolor}
\usepackage[linesnumbered,ruled,vlined]{algorithm2e}
\graphicspath{{./img/}}
\DeclareGraphicsExtensions{.png,.jpg,.eps,.pdf}
\IEEEoverridecommandlockouts

\author{Boyu Zhou, Fei Gao, Luqi Wang, Chuhao Liu and Shaojie Shen%
\thanks{Accepted version. To appear in the IEEE Robotics and Automation Letters. \textcircled{\small{c}}2019 IEEE.  Personal use of this material is permitted. Permission from IEEE must be obtained for all other uses.
This work was supported by the Robotics Institue, Hong Kong University of Science and Technology.} 
\thanks{All authors are with the Department of Electronic and Computer Engineering, Hong Kong University of Science and Technology, Hong Kong, China. {\tt\footnotesize $\{$bzhouai, fgaoaa, lwangax, cliuci, eeshaojie$\}$@ust.hk}}%
\thanks{Digital Object Identifier (DOI): see top of this page.}
}

\title{Robust and Efficient Quadrotor Trajectory Generation for Fast Autonomous Flight}

\begin{document}

\maketitle

\setstcolor{blue}

\begin{abstract}
In this paper, we propose a robust and efficient quadrotor motion planning system for fast flight in 3-D complex environments. We adopt a kinodynamic path searching method to find a safe, kinodynamic feasible and minimum-time initial trajectory in the discretized control space. {We improve the smoothness and clearance of the trajectory by a B-spline optimization, which incorporates gradient information from a Euclidean distance field (EDF) and dynamic constraints efficiently utilizing the convex hull property of B-spline. Finally, by representing the final trajectory as a non-uniform B-spline, an iterative time adjustment method is adopted to guarantee dynamically feasible and non-conservative trajectories.} We validate our proposed method in various complex simulational environments. The competence of the method is also validated in challenging real-world tasks. {We release our code as an open-source package}\footnote{Open-source implementation is available at \url{https://github.com/HKUST-Aerial-Robotics/Fast-Planner}.}.
\end{abstract}


\IEEEpeerreviewmaketitle

\section{Introduction}
\IEEEPARstart{U}{nmanned} aerial vehicles (UAVs) are recently involved in more and more applications, such as industrial inspection, search-and-rescue and package delivery. To achieve full autonomy in these scenarios, the motion planning module plays an essential role in generating safe and smooth motions. 

Although plenty of works on quadrotor trajectory generation have been proposed, there are still two critical unsolved issues. Firstly, given limited time and onboard computing resources, no existing works guarantee to generate safe and kinodynamic feasible trajectory at a high success rate. However, the efficiency and robustness of the trajectory generation are essential. In many circumstances, such as a quadrotor flying at high speed in unknown environments, trajectories should be re-generated constantly in a very short time to avoid emergent threats. Secondly, to ensure the kinodynamic feasibility of the generated motions, constraints on velocity and acceleration are often enforced conservatively. As a result, the aggressiveness of the generated trajectories are often hard to be tuned to satisfy applications where a high flight speed is preferable.

\begin{figure}[t]
	\begin{center}          
	\subfigure[\label{fig:bag1} Fast autonomous flight in an unknown environment.]
	{\includegraphics[width=0.48\columnwidth]{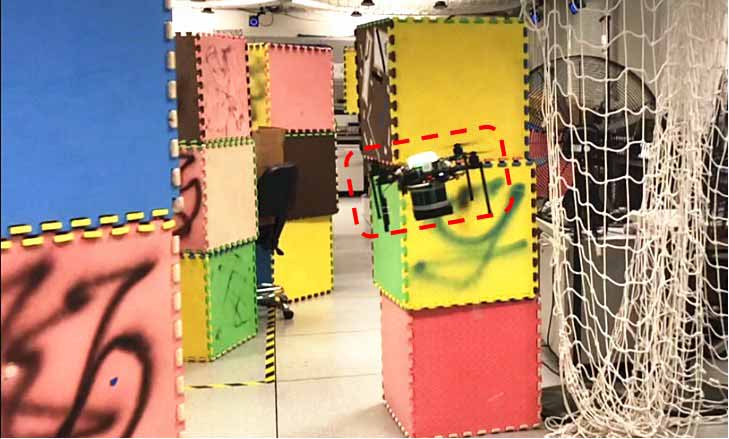}}       
	\subfigure[\label{fig:bag2} Aggressive kinodynamic replans.]
	{\includegraphics[width=0.48\columnwidth]{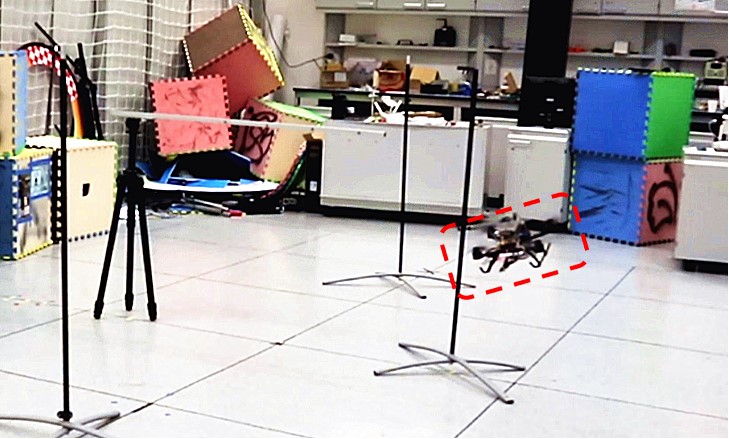}}     
	\end{center}
	\caption{\label{fig:indoor_bag}Our proposed method tested on a fully autonomous quadrotor in (a), and on extremely challenging fast replanning with indoor external feedback in (b). Experimental details are given in Sect.~\ref{sect:res}. Video is available at \url{https://www.youtube.com/watch?v=GIYGAjOeeI8&feature=youtu.be} }
\end{figure}  

In this paper, we propose a complete and robust online trajectory generation method to address these two issues systematically. A kinodynamic path searching based on {heuristic search} and linear quadratic minimum-time control is adopted. It searches efficiently for a safe, feasible and minimum-time initial path in the discretized control space. 
{The initial path is then refined in a carefully designed B-spline optimization, which utilizes B-spline's convex hull property to incorporate gradient information and dynamic constraints. It improves the initial path and converges quickly to a smooth, safe and dynamically feasible trajectory. 
Finally, the trajectory is represented as a non-uniform B-spline, for which we investigate the relations between the control points of derivatives and time allocation. Based on the relations, an iterative time adjustment method is adopted to squeeze infeasible velocity and acceleration out from the profiles while avoiding constraining them conservatively.}

Compared to existing works, our proposed method is able to generate high-quality trajectories in cluttered environments in a much shorter time with a higher success rate. {It can generate aggressive motion under the premise of dynamic feasibility.} We show the efficiency and robustness of our method in numerous simulational complex environments. We also demonstrate that our method is competent even for challenging fast flight when trajectories should be re-generated repeatedly in a very short time by real-world experiments. We summarize our contributions as follows:

1) We propose a robust and efficient systematic method, incorporating kinodynamic path searching, B-spline optimization and time adjustment, where safety, dynamic feasibility and aggressiveness are built from bottom-up.

2) {We present an optimization formulation based on the convex hull property of B-splines that delicately incorporates gradient information and dynamic constraints, which converges quickly to generate smooth, safe and dynamically feasible trajectories. 

3) We investigate the relations between the control points of derivatives and the time allocation of non-uniform B-splines. 
A time adjustment method based on the relations is applied to guarantee feasible and non-conservative motion.}

4) We present extensive simulation and real-world evaluation of our proposed method. The source code is released as a ros-package.

\vspace{-0.3cm}
\section{Related Work}
\label{sect:rel}
\subsubsection{Hard-constrained methods}
The problems of trajectory generation have been addressed by some work recently. Hard-constrained methods are pioneered by minimum-snap trajectory generation~\cite{MelKum1105}, in which trajectories are represented as piecewise polynomials and generated by solving a quadratic programming(QP) problem. \cite{RicBryRoy1312} shows that minimum snap trajectories can be obtained in closed form, in which the safety of the trajectories is ensured by iteratively adding intermediate waypoints. Works\cite{CheSuShe2015,fei2016ssrr,liu2017planning, ding2018trajectory, ding2019efficient} generate trajectories in a two-step pipeline. Free space represented by a sequence of cubes\cite{CheSuShe2015, fei2018icra}, spheres\cite{fei2016ssrr,fei2018jfr} or polyhedrons\cite{liu2017planning} is firstly extracted, which is followed by convex optimization, which generates smooth trajectory within the feasible space. \cite{ding2018trajectory,ding2019efficient} proposed a B-spline-based kinodynamic search to find an initial trajectory which is then refined by an elastic band optimization approach. The use of uniform B-spline ensures dynamic feasibility but could generate conservative motion. One common drawback of these methods is that the time allocation of the trajectory is given by naive heuristics. However, a poorly chosen time allocation significantly reduce the quality of the trajectory. Besides, a feasible solution can only be obtained by iteratively adding more constraints and solving the quadratic programming problem, which is undesirable for real-time application. To address these problems, \cite{fei2018icra} proposed a method to search for a path with well-allocated time and guarantee the safety and kinodynamic feasibility of trajectory through optimization. Hard-constrained methods ensure global optimality by the convex formulation. However, distance to obstacles in the free space is ignored, which often results in trajectories being close to obstacles. Besides, the kinodynamic constraints are conservative, making the trajectory's speed deficient for fast flight.

\subsubsection{Soft-constrained methods}
There are also methods formulating trajectory generation as a non-linear optimization problem that takes smoothness and safety into account. \cite{zucker2013chomp} generates discrete-time trajectories by minimizing its smoothness and collision costs using gradient descent methods.\cite{kalakrishnan2011stomp} has similar problem formulation, but the optimization is solved by a gradient-free sampling method. \cite{oleynikova2016continuous} extended them to continuous-time polynomial trajectories. Since the time parameterization is continuous, it avoids numeric differentiation errors and is more accurate to represent the motions of quadrotors. However, it suffers from a low success rate. To solve this problem, \cite{fei2017iros} finds a collision-free initial path firstly using an informed sampling-based path searching method. This path serves as a higher quality initial guess of non-linear optimization and thus improve the success rate. In \cite{usenko2017real}, the trajectory is parameterized as a uniform B-spline. Since a B-spline is continuous by nature, there is no need to enforce continuity explicitly, which reduce the number of constraints. It is also particularly useful for local replanning thanks to its property of locality. Soft-constrained methods utilize gradient information to push trajectory far from obstacles, but suffer from local minima and having no strong guarantee of success rate and kinodynamic feasibility. {Our optimization method also utilizes gradient information to improve the safety of the trajectory. However, unlike previous methods in which computational expensive line integrals along the trajectory are calculated, the formulation is redesigned to be simpler based on the convex hull property of B-spline. It greatly improves the computation efficiency as well as the convergent rate.} 

\begin{figure}[t]
	\centering
	\includegraphics[width=0.8\columnwidth]{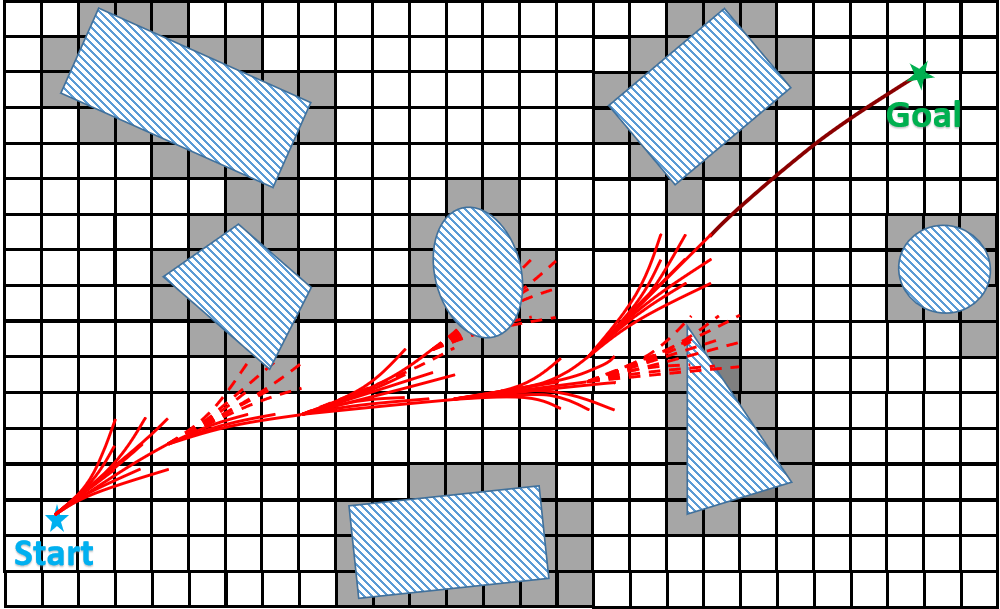} 
	\caption{An illustration of the mechanism of the kinodynamic path searching. {Red curves indicate the motion primitives generated by Equ.\ref{equ:solution}}. The purple curve is the analytic expansion explained in Sect.~\ref{subsect:shooting_expansion}.} 
	\label{img:search}
	\vspace{-0.6cm}
\end{figure}
\section{Kinodynamic Path Searching}
\label{sect:a*}
Our front-end kinodynamic path searching module is originated from the hybrid-state A* search first proposed for autonomous vehicle\cite{dolgov2010path}. It searches for a safe and kinodynamic feasible trajectory that is minimal with respect to time duration and control cost in a voxel grid map. 
{
As shown in Alg.\ref{alg:search} and in Fig.\ref{img:search}, the searching loop is similar to the standard A* algorithm, where $ \mathcal{P} $ and $ \mathcal{C} $ refer to the open and closed set. 
Instead of straight lines, motion primitives respecting the quadrotor dynamic are used as graph edge. A structure \textit{Node} is used to record a primitive, the voxel the primitives ends in and the $g_c$ and $ f_c $ cost (Sect. \ref{subs:cost}). 
Primitives \textbf{Expand}() the voxel grid map iteratively and those ending up in the same voxel except the one with the smallest $ f_c $ are pruned (\textbf{Prune}()). 
Then \textbf{CheckFeasible}() checks the safety and dynamic feasibility of the remained primitives. This process continues until any primitive reach goal or the \textbf{AnalyticExpand()} (Sect. \ref{subsect:shooting_expansion}) succeeds.}
%
\begin{algorithm}
	Initialize()\;
	{
	\While{$\lnot \ \mathcal{P}.\textnormal{\textbf{empty}}()$}{
		$ n_c \gets \mathcal{P}$.\textbf{pop}(), $\mathcal{C} $.\textbf{insert}($ n_c $) \;
		\If{$ \textnormal{\textbf{ReachGoal}}(n_c) \lor \textnormal{\textbf{AnalyticExpand}}(n_c)$}{
			\Return{\textnormal{\textbf{RetrievePath}()}}\;
		}
		$ primitives \gets $ \textbf{Expand}($ n_c $)\; 
		$ nodes \gets \textnormal{\textbf{Prune}}( primitives ) $\; 
		\For{$ n_i \ \textbf{in} \ nodes $}{
			\If{$ \lnot \ \mathcal{C}.\textnormal{\textbf{contain}}(n_i) \land \textnormal{\textbf{CheckFeasible}}(n_i)  $}{
				$ g_{temp} \gets n_c.g_c + \textnormal{\textbf{EdgeCost}}(n_i)$ \;
				\If{$ \lnot \ \mathcal{P}.\textnormal{\textbf{contain}}(n_i) $}{
					$ \mathcal{P} $.\textbf{add}($ n_i $)\;						
				}
				\ElseIf{$ g_{temp} \geq n_i.g_c $}{
						continue\;
				}
				$ n_i.parent \gets n_c, \ n_i.g_c \gets g_{temp} $\;
				$ n_i.f_c \gets n_i.g_c$ + \textbf{Heuristic}($ n_i$)\;
			}
		}
	}
}
\caption{Kinodynamic Path Searching \label{alg:search}}
\end{algorithm}
\subsection{Primitives Generation}
\label{subs:chil}
{We first discuss the generation of motion primitives used in \textbf{Expand}().}
The differential flatness of quadrotor systems allows us to represent the trajectory by {three independent 1-D time-parameterized polynomial functions\cite{MelKum1105}:
\begin{equation}
	\label{equ:poly}
	\mathbf{p}(t):=\left[ p_{x}(t), p_{y}(t), p_{z}(t)  \right]^{\top} ,\quad p_{\mu}(t) = \sum_{k=0}^{K} a_{k} t^{k}
\end{equation}
where $ \mu \in \{x,y,z\} $. 
From the view of quadrotor systems, it corresponds to a linear time-invariant (LTI) system. Let $ \mathbf{x}(t):= [\mathbf{p}(t)^{\top}, \dot{\mathbf{p}}(t)^{\top}, \cdots, \mathbf{p}^{(n-1)}(t)^{\top}]^{\top} \in \mathcal{X} \subset \mathbb{R}^{3 n} $ be the state vector. Let $ \mathbf{u}(t) := \mathbf{p}^{(n)}(t) \in \mathcal{U} := \left[-u_{\max }, u_{\max }\right]^{3} \subset \mathbb{R}^{3} $ be the control input. The state space model can be defined as:}  
\begin{align}\label{equ:state_space}
	\dot{\mathbf{x}} &= \mathbf{A}\mathbf{x}+\mathbf{B}\mathbf{u} \\ \nonumber
	\mathbf{A} &= \left[ \begin{array}{ccccc}
	\mathbf{0} & \mathbf{I}_{3} & \mathbf{0} & \cdots & \mathbf{0} \\
	\mathbf{0} & \mathbf{0} &  \mathbf{I}_{3} & \cdots & \mathbf{0} \\
	\vdots & \vdots & \vdots &  \ddots & \vdots \\
	\mathbf{0} & \cdots & \cdots & \mathbf{0} &  \mathbf{I}_{3} \\
	\mathbf{0} & \cdots & \cdots & \mathbf{0} & \mathbf{0}
	\end{array} \right], 
	\quad \mathbf{B} = \left[ \begin{array}{c}
	\mathbf{0} \\ \mathbf{0} \\ \vdots \\ \mathbf{0} \\ \mathbf{I}_{3}	
	\end{array} \right]
\end{align}
{The complete solution for the state equation is expressed as:}
\begin{equation}\label{equ:solution}
	\mathbf{x}(t) = e^{\mathbf{A}t}\mathbf{x}(0) + 
	\int_{0}^{t}e^{\mathbf{A}(t-\tau)}\mathbf{B}\mathbf{u}(\tau) \ d\tau
\end{equation}
{which gives the trajectory of the quadrotor system whose initial state is $ \mathbf{x}(0) $ and control input
is $ \mathbf{u}(t) $.

In \textbf{Expand}(), given the current state of the quadrotor, a set of discretized control inputs $ \mathcal{U}_D \subset \mathcal{U} $ is applied for duration $ \tau $. In practice, we select $ n = 2 $, which corresponds to a double integrator. Each axis $ \left[ -u_{max}, u_{max} \right] $ is discretized uniformly as $ \{-u_{max}, -\frac{r-1}{r}u_{max}, \cdots, \frac{r-1}{r}u_{max}, \ u_{max}\} $, which results in $ (2r+1)^3 $ primitives.}
\subsection{Actual Cost and Heuristic Cost}
\label{subs:cost}
As we aim to find a trajectory that is optimal in time and control cost, we define the cost of a trajectory as:
\begin{equation}\label{equ:cost_search}
	\mathcal{J}(T) = \int_{0}^{T} \Vert \mathbf{u}(t) \Vert^{2} dt + \rho T
\end{equation}
{Under this definition, \textbf{EdgeCost}() calculates the cost of a motion primitive generated with the discretized input $ \mathbf{u}(t) = \mathbf{u}_d $ and duration $ \tau $ as $ e_c = (\Vert \mathbf{u}_{d} \Vert^{2} + \rho)\tau $. 

Following the terminology of A*, we use $ g_c $ to represent the actual cost of an optimal path from the start state $ \mathbf{x}_s $ to the current state $ \mathbf{x}_c $. Let this optimal path consists of $ J $ primitives, $ g_c $ is calculated as: $ g_c = \sum_{j=1}^{J} (\Vert \mathbf{u}_{dj} \Vert^{2} + \rho)\tau $.}

An admissible and informative heuristic is essential to speed up the searching as in A*. Hence, we also design a {\textbf{Heuristic}().
We compute a closed form trajectory that minimizes $ \mathcal{J}(T) $ from $ \mathbf{x}_c $ to the goal state $ \mathbf{x}_g $ by applying the Pontryagin’s minimum principle\cite{mueller2015computationally}:
\begin{align}\label{equ:poly}
	p_{\mu}^{*}(t) &= \frac{1}{6}\alpha_{\mu} t^{3} + \frac{1}{2}\beta_{\mu} t^{2} + v_{\mu c} + p_{\mu c} \\ \nonumber
\left[ \begin{array}{c}{\alpha_{\mu}} \\ {\beta_{\mu}} \end{array}\right] &= \frac{1}{T^{3}} \left[ \begin{array}{cc}{-12} & {6 T} \\ {6 T} & {-2 T^{2}}\end{array}\right] \left[ \begin{array}{c}{p_{\mu g}-p_{\mu c}-v_{\mu c} T} \\ {v_{\mu g}-v_{\mu c}}\end{array}\right] \\ \nonumber
\mathcal{J}^{*}(T) &= \sum_{\mu \in \{x,y,z\}}(\frac{1}{3}\alpha_{\mu}^{2}T^{3} + \alpha_{\mu} \beta_{\mu} T^{2} + \beta_{\mu}^{2} T)
\end{align} 
where $ p_{\mu c}, v_{\mu c}, p_{\mu g}, v_{\mu g} $ are the current and goal position and velocity. $ \mathcal{J}^{*}(T) $ is the cost defined by Equ.\ref{equ:cost_search}. To find the optimal time $ T $ that minimize the cost, we substitute $ \alpha_{\mu}, \beta_{\mu} $ into $ \mathcal{J}^{*}(T) $ and find the roots of $ \frac{\partial \mathcal{J}^{*}(T)}{\partial T} = 0 $.} The root making a minimum cost $ min \ \mathcal{J}^{*} $ and feasible trajectory is selected {and denoted as $ T_{h} $. We use $ \mathcal{J}^{*}(T_{h}) $ as the heuristic $ h_c $. Finally, $ f_c $ is defined as:
$	f_c = g_c + h_c = g_c + \mathcal{J}^{*}(T_{h}) $.}


\subsection{Analytic Expansion}
\label{subsect:shooting_expansion}
Due to the discretized control input, it is difficult to have a primitive end exactly in the goal state. To compensate for it and also to speed up the searching, we induce an analytic expansion scheme. {When a node is popped from the open set, a trajectory from $ \mathbf{x}_c $ to $ \mathbf{x}_g $ is computed using the same approach in Sect. \ref{subs:cost}. If it passes the safety and dynamic feasibility check, the searching is terminated in advance. This strategy is effective for improving efficiency especially in sparse environments, since it has a higher success rate and terminates the searching earlier.}

{
\subsection{Optimality and Completeness}
\label{subsect:optimality}
Theoretically, we can not guarantee the optimality and completeness of the path searching. 
However, the practical performance is satisfactory. For the optimality, evaluation (Sect. \ref{ssubs:analy}) shows that the sacrifice of optimality is acceptable and adjustable. Besides, provided the initial path lies near the optimum, our optimization (Sect. \ref{sect:opt}) will find that optimum.
For the completeness, evaluation (Sect. \ref{ssubs:analy}) indicates that in practice it can find a feasible solution in most case. Also, our method can be extended to support variable-duration primitives and a variable-resolution voxel grid as described by \cite{dolgov2010path} to make stronger completeness guarantees.
}

\section{B-spline Trajectory Optimization}
\label{sect:opt}
\begin{figure}[t]
	\centering
	\includegraphics[width=0.98\columnwidth]{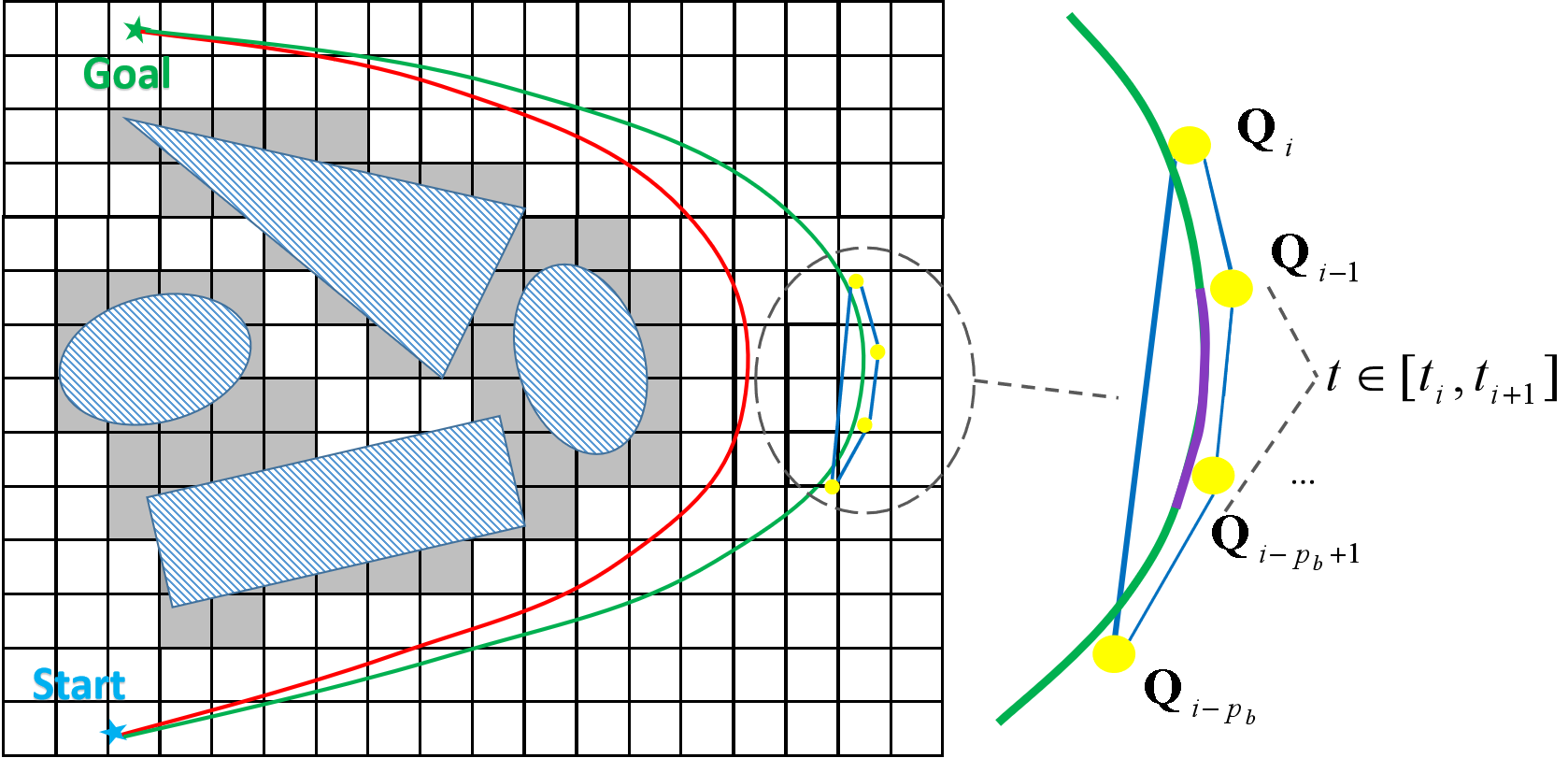} 
	\caption{Using gradient-based numeric optimization to deform the trajectory. The red and the green curves are the initial path and the B-spline after the optimization. Yellow dots stand for the control points of the B-spline. {The initial path is close to the obstacles since distance information is ignored, while the B-spline is pushed away by the gradient-based optimization.} \label{fig:optimization}} 
\end{figure}

{As mentioned in Sect. \ref{subsect:optimality}, the path produced by the path searching can be suboptimal. In addition, this path is often close to obstacle since distance information in the free space is ignored (Fig. \ref{fig:optimization}). Therefore, we improve the smoothness and clearance of the path in the proposed B-spline optimization. The \textbf{convex hull} property of uniform B-splines are utilized to incorporate gradient information from the Euclidean distance field and dynamic constraints, for which it converges within a very short duration to generate smooth, safe and dynamically feasible trajectories. }
\subsection{Uniform B-splines}
\label{subs:spline}
A B-spline is a piecewise polynomial uniquely determined by its degree {$ p_b $}, a set of $ N+1 $ {\textbf{control points}} $ \{\mathbf{Q}_{0},\mathbf{Q}_{1}, \cdots, \mathbf{Q}_{N}  \} $ and a {\textbf{knot vector}} $ [ t_{0}, t_{1}, \cdots, {t_{M}} ] $, in which $ \mathbf{Q}_{i} \in \mathbb{R}^{3} $, $ {t_{m}} \in \mathbb{R} $ and $ M = N+{p_b}+1 $. {A B-spline trajectory is parameterized by time $ t $, where $ t \in \left[ t_{p_b}, t_{M-p_b} \right] $. For a \textbf{uniform} B-spline, each \textbf{knot span} $ \Delta t_{m} = t_{m+1} - t_{m} $ has identical value $ \Delta t $. To evaluate the position at time $ t \in [t_{m}, t_{m+1}) \subset \left[ t_{p_b}, t_{M-p_b} \right] $, we first normalize $ t $ as $ s(t) = (t - t_{m})/ \Delta t$. Then the position can be evaluated using the matrix representation\cite{qin2000general}}:
{
\begin{align}\label{equ:matrix}
	& \mathbf{p}(s(t)) = \mathbf{s}(t)^{\top} \mathbf{M}_{p_b+1} \mathbf{q}_m \\ \nonumber
	& \mathbf{s}(t) = \left[
		\begin{matrix}
		1 & s(t) & s^{2}(t) & \cdots & s^{p_b}(t) 
		\end{matrix}
	\right]^{\top} \\ \nonumber 
	& \mathbf{q}_m = \left[
		\begin{matrix}
		\mathbf{Q}_{m-p_b} & \mathbf{Q}_{m-p_b+1}& \mathbf{Q}_{m-p_b+2} & \cdots & \mathbf{Q}_{m}
		\end{matrix}
	\right]^{\top}
\end{align}
here $ \mathbf{M}_{p_b+1} $ is a constant matrix determined by $ p_b $. In our implementation, $ p_b $ is set as $ 3 $. The evaluation of the derivatives is exactly the same, since the derivative of a B-spline is also a B-spline.}

{The \textbf{convex hull} property of B-splines (Fig.\ref{fig:convex}) is essential for designing our optimization formulation. We show in Sect.\ref{subs:convex_hull} that it is extremely useful for ensuring the dynamic feasibility and safety of the entire trajectory. }	

{
\subsection{Convex Hull Property}
\label{subs:convex_hull}
\begin{figure}[t]
	\begin{center}          
		\subfigure[]
		{\includegraphics[width=0.56\columnwidth]{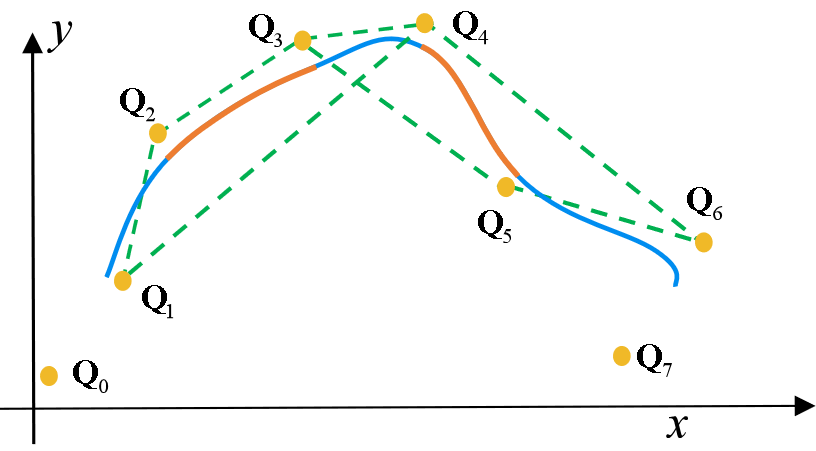}}       
		\subfigure[]
		{\includegraphics[width=0.34\columnwidth]{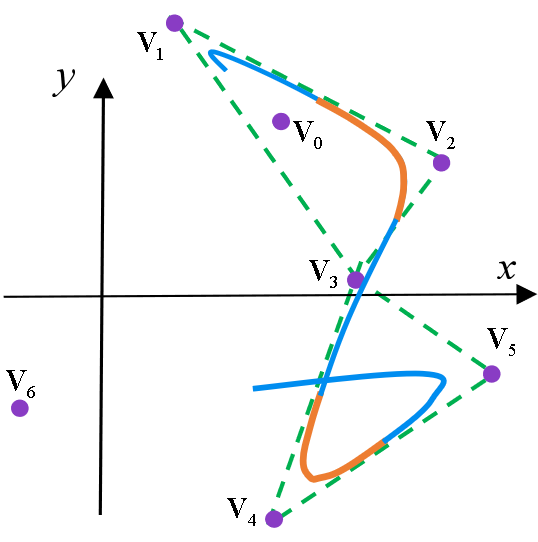}}     
	\end{center}
	\caption{\label{fig:convex} {a) A trajectory is represented by a B-spline ($ p_b = 3 $). Each segment is bounded by the corresponding convex hull of the control points (example convex hulls and segments are shown in green and orange). b) The first order derivative (velocity) is also a B-spline, thus it has the same property. The control points of the derivatives can be calculated by Equ.\ref{equ:con}.}  }
\vspace{-0.3cm}
\end{figure}  
\begin{figure}[t]
	\begin{center}          
		\includegraphics[width=0.7\columnwidth]{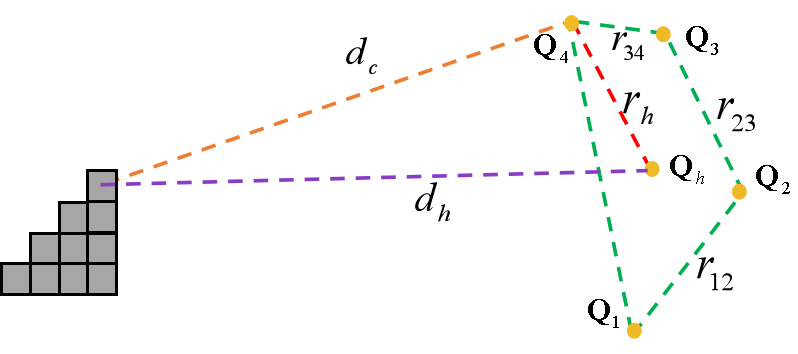}     
	\end{center}
	\caption{\label{fig:convex2} {Illustration of ensuring that a convex hull of the B-spline ($ p_b = 3 $) is collision-free.}}
	\vspace{-0.7cm}
\end{figure}    
The convex hull property (Fig.\ref{fig:convex}) is used extensively in our method to ensure both dynamic feasibility and safety. 

For the dynamic feasibility, it suffices to constrain all velocity and acceleration control points $ \{\mathbf{V}_{0}, \mathbf{V}_{1}, \cdots, \mathbf{V}_{N-1} \} $ and $ \{\mathbf{A}_{0}, \mathbf{A}_{1}, \cdots, \mathbf{A}_{N-2} \} $ so that $ \mathbf{V}_{i} \in \left[-v_{\max }, v_{\max }\right]^{3} $ and $ \mathbf{A}_{i} \in \left[-a_{\max }, a_{\max }\right]^{3} $. $ \mathbf{V}_{i} $ and $ \mathbf{A}_{i} $ are calculated by Equ.\ref{equ:con}, where $ \Delta t $ is the knot span:
\begin{equation}\label{equ:con}
	\mathbf{V}_{i} = \frac{1}{\Delta t}(\mathbf{Q}_{i+1}-\mathbf{Q}_{i}), \quad \mathbf{A}_{i} = \frac{1}{\Delta t}(\mathbf{V}_{i+1}-\mathbf{V}_{i}) 
	\end{equation}
}
{
For the safety of the B-spline, we need to ensure that all its convex hulls are collision-free. 
Equivalently, we need to ensure that $ d_{h} > 0$, where $ d_{h} $ is the distance between any one occupied voxel and any one point $ Q_{h} $ in the convex hull (Fig.\ref{fig:convex2}). By the triangle inequality, we have $ d_h > d_{c} - r_{h} $, where $ d_c $ is the distance between the voxel and any one control point. We also have $ r_h \le r_{12} + r_{23} + r_{34}$, since $ Q_h $ is inside the convex hull. Combining them, $ d_h > d_{c} - (r_{12} + r_{23} + r_{34}) $ is always valid.
Therefore, if we ensure:
\begin{equation}
\label{equ:safety}
	d_{c} > 0, \quad r_{j,j+1} < d_{c}/3 \ \ (j \in \{1,2,3 \})
\end{equation}  
then the convex hull is guaranteed to be collision-free.
}
%
\subsection{Problem Formulation}\label{subs:formulation}

For a $ p_b $ degree B-spline trajectory defined by $ N+1 $ control points $ \{\mathbf{Q}_{0},\mathbf{Q}_{1}, \cdots, \mathbf{Q}_{N}\} $, we optimize the subset of {$ N+1-2{p_b} $} control points {$ \{\mathbf{Q}_{p_b},\mathbf{Q}_{p_b+1}, \cdots, \mathbf{Q}_{N-p_b}\} $}. The first and last {$ p_b $} control points {should not be changed because they determine the boundary state}. The overall cost function is defined as:
\begin{equation}\label{equ:cost}
	f_{total} = \lambda_{1} f_{s} + \lambda_{2} f_{c} + \lambda_{3} (f_{v} + f_{a})
\end{equation}
where $ f_{s} $ and $ f_{c} $ are smoothness and collision cost. $ f_{v} $ and $ f_{a} $ are soft limits on velocity and acceleration. {$ \lambda_1, \lambda_2 $ and $ \lambda_3 $ trade off the smoothness, safety and dynamic feasibility.}

We define the smoothness cost $ f_{s} $ by a function that captures the geometric information of the trajectory and does not depend on time allocation, unlike many recent works that use integral of the squared snap or jerk. The reason is that after optimization the time allocation may be adjusted (Sect. \ref{sect:knot}). This will change the derivatives of the trajectory and make the optimized {snap (jerk)} less meaningful. We use an elastic band cost function \footnote{{The control points $ \mathbf{Q}_{p_b-2}, \mathbf{Q}_{p_b-1}, \mathbf{Q}_{N-p_b+1} $ and $ \mathbf{Q}_{N-p_b+2} $ are not optimized but needed to evaluate the overall smoothness. Similarly, some of them are included in Equ.\ref{equ:kino} as constant values to evaluate the derivatives.}} \cite{quinlan1993elastic,zhu2015convex}:
\begin{equation}\label{equ:elastic} 
	f_{s} 
	= \sum\limits_{i=p_b-1}^{N-p_b+1} \Vert \underbrace{(\mathbf{Q}_{i+1}-\mathbf{Q}_{i})}_{\mathbf{F}_{i+1,i}} \ + \ \underbrace{(\mathbf{Q}_{i-1}-\mathbf{Q}_{i})}_{\mathbf{F}_{i-1,i}} \Vert^{2} 
\end{equation}
From a physical standpoint, this formulation view a trajectory as an elastic band, where each term {$ \mathbf{F}_{i+1,i}  = \mathbf{Q}_{i+1} -\mathbf{Q}_{i} $ and $ \mathbf{F}_{i-1,i} = \mathbf{Q}_{i-1}-\mathbf{Q}_{i}$} is the joint force of two springs connecting the nodes $ \mathbf{Q}_{i+1}, \mathbf{Q}_{i} $ and $ \mathbf{Q}_{i-1}, \mathbf{Q}_{i} $ respectively. If all terms equal to zero, all the control points would uniformly distribute in a straight line, which is ideally smooth. 

Similarly, the collision cost is formulated as the repulsive force of the obstacles acting on each control point:
\begin{equation}\label{equ:colli}
	f_{c} = \sum\limits_{i=p_b}^{N-p_b} F_{c}(d(\mathbf{Q}_{i}))
\end{equation}
{where $ d(\mathbf{Q}_{i}) $ is the distance between $ \mathbf{Q}_{i} $ and the closet obstacle.} $ F_{c} $ is a differentiable potential cost function with {$ d_{thr} $} specifying the threshold of obstacle clearance:
{
\begin{equation}\label{equ:potential}
	F_{c}(d(\mathbf{Q}_{i})) = \left\{
	\begin{array}{cl}
	(d(\mathbf{Q}_{i})-d_{thr})^{2} & d(\mathbf{Q}_{i}) \le d_{thr} \\
	0 & d(\mathbf{Q}_{i}) > d_{thr}
	\end{array}
	 \right.
\end{equation}
}

{	
As shown in Sect.\ref{subs:convex_hull}, Equ.\ref{equ:safety} must be satisfied so that the trajectory is collision-free. Since the collision cost pushes the control points away from obstacles, $ d_{c} > 0 $ is apparent. Also, $ r_{j,j+1} $ are tunable parameters depend solely on the parameterization of the B-spline. 
In practice, as long as we select $ r_{j,j+1}, (j \in \{0,1,\cdots,N \}) $ that are significantly small (in our implementation $ r_{j,j+1} < 0.2 $), the trajectory is safe in most cases. This may be invalid in extreme cases, for instance, the environment is very cluttered. Even so, we can re-parameterize the B-spline to select smaller $ r_{j,j+1} $, after which Equ.\ref{equ:safety} will still be satisfied.
}

We penalize velocity or acceleration along the trajectory exceeding maximum allowable value {$v_{max}$ and $ a_{max} $ } with a cost similar to Equ.\ref{equ:potential}. The penalty for {1-D velocity $ v_{\mu} $} is:
\begin{equation}\label{key}
	F_{v}(v_{\mu}) = \left\{
	\begin{array}{ccl}
	(v_{\mu}^{2} - v_{max}^{2})^{2} & & v_{\mu}^{2} > v_{max}^{2} \\ 
	0 & & v_{\mu}^{2} \le v_{max}^{2}
	\end{array}
	\right.
\end{equation}
{where $ \mu \in \{x,y,z \} $. The acceleration penalty has identical form. Applying the convex hull property (Fig. \ref{fig:convex}), we define $ f_{v} $ and $ f_{a} $ so that infeasible velocity and acceleration control points are penalized:}
{
\begin{equation}\label{equ:kino}
	f_{v} = \sum\limits_{\begin{subarray}{c}
		\mu \in \\ \{x,y,z \}
	\end{subarray}} \sum\limits_{i=p_b-1}^{N-p_b} F_{v}(V_{i\mu}), \quad
	f_{a} = \sum\limits_{\begin{subarray}{c}
		\mu \in \\ \{x,y,z \}
	\end{subarray}} \sum\limits_{i=p_b-2}^{N-p_b} F_{a}(A_{i\mu})
\end{equation}
}

\vspace{-1cm}
\section{Time Adjustment}
\label{sect:knot}
Although we constrain kinodynamic feasibility in the path searching and optimization, sometimes we get infeasible trajectories. The basic reason is that gradient information tends to lengthen the overall trajectory while pushing it far from obstacles. Consequently, the quadrotor has to fly more aggressively in order to travel longer distance within the same time, which unavoidably causes over aggressive motion if the original motion is already near to the physical limits. 

{To guarantee dynamic feasibility, we adopt a time adjustment method based on the relations between the derivatives control points and the time allocation (knot spans) of the non-uniform B-spline. Thanks to the relations, we can change the flight aggressiveness as we expected by adjusting the associated time allocation. Thus dynamic feasibility can be ensured without over-conservative constraints.

We first introduce the mathematic fundament of the time adjustment. Then the Alg. \ref{alg:adj} is presented to tackle over-aggressive trajectories.}

\subsection{Non-uniform B-spline}
\label{subs:non}
Non-uniform B-spline is a more general kind of B-spline. The only difference to uniform B-spline is that each of its knot span {$ \Delta t_{m} = t_{m+1} - t_{m} $} is independent to others. The control points of a non-uniform B-spline's first and second order derivatives $ \mathbf{V}_{i}^{'} $ and $ \mathbf{A}_{i}^{'} $ can be computed by:
{
\begin{equation}\label{equ:con-non}
	\mathbf{V}_{i}^{'} = \frac{p_b(\mathbf{Q}_{i+1}-\mathbf{Q}_{i})}{t_{i+p_b+1}-t_{i+1}}, \ \ \mathbf{A}_{i}^{'} = \frac{(p_b-1)(\mathbf{V}_{i+1}^{'}-\mathbf{V}_{i}^{'})}{t_{i+p_b+1}-t_{i+2}}
\end{equation}
}
{
By the convex hull property (Fig.\ref{fig:convex}), to enforce the dynamic feasibility of a trajectory represented by a non-uniform B-spline, it suffices to enforce all control points of the first and second order derivatives within the feasible domain. We show that this can be achieved by changing the corresponding knot spans of the infeasible control points in Sect. \ref{subs:adj}.}

\subsection{Knot Spans Adjustment}
\label{subs:adj}
{
Let $ \mathbf{V}_{i}^{'} = [ V_{i,x}^{'}, V_{i,y}^{'}, V_{i,z}^{'} ]^{\top} $ be an infeasible control point of velocity. Let $ V_{i,\mu}^{'} $ be the largest infeasible component and $ \mid V_{i,\mu}^{'} \mid = v_m $. From Equ.\ref{equ:con-non} we know that $ V_{i,\mu}^{'} $ is influenced by the duration $ t_{i+p_b+1} - t_{i+1} $. If we change this duration to $ \hat{t}_{i+p_b+1} - \hat{t}_{i+1} = \mu_v (t_{i+p_b+1} - t_{i+1}) $, then $ V_{i,\mu}^{'} $ changes to:
\begin{align}
\label{equ:prove_vel}
 \hat{V}_{i,\mu} &= \frac{p_b}{\hat{t}_{i+p_b+1}-\hat{t}_{i+1}} ( Q_{i+1,\mu}-Q_{i,\mu} ) \\ \nonumber
	&= \frac{1}{\mu_{v}} \frac{p_b}{t_{i+p_b+1}-t_{i+1}} ( Q_{i+1,\mu}-Q_{i,\mu} ) = \frac{1}{\mu_{v}} {V}_{i,\mu}^{'} 
\end{align}
Therefore, if we set $ \mu_{v} = \frac{v_{m}}{v_{max}} $, then the velocity is feasible, because $ \mid \hat{V}_{i,\mu} \mid = \frac{v_{max}}{v_{m}} \mid {V}_{i,\mu}^{'} \mid = v_{max} \in \left[ -v_{max}, v_{max} \right] $.

The enforcement of acceleration feasibility is similar\footnote{{We use the same notation as those in the last paragraph and do not define them explicitly for brevity.} }. We know that $ A_{i,\mu}^{'} $ is actually influenced by $ t_{i+p_b+2} - t_{i+1} $ since it is coupled with $ V_{i,\mu}^{'} $ and $ V_{i+1,\mu}^{'} $. We change $ \Delta t_m = t_{m+1} - t_{m} $ to $ \Delta \hat{t}_{m} = \mu_a \Delta t_m $ for $ m \in \{i+1,i+2, \cdots, i+p_b+1 \} $ and we get:
\begin{align}
\label{equ:prove_acc}
\hat{A}_{i,\mu} &= \frac{p_b-1}{\hat{t}_{i+p_b+1}-\hat{t}_{i+2}} ( \hat{V}_{i+1,\mu}- \hat{V}_{i,\mu} ) \\ \nonumber
& = \frac{1}{\mu_a} \frac{p_b-1}{t_{i+p_b+1}-t_{i+2}} ( \frac{1}{\mu_a} V_{i+1,\mu}^{'} - \frac{1}{\mu_a} V_{i,\mu}^{'} ) \\ \nonumber
& =  \frac{1}{\mu_{a}^{2}} \frac{p_b-1}{{t}_{i+p_b+1}-{t}_{i+2}} ( {V}_{i+1,\mu}^{'} - {V}_{i,\mu}^{'} ) = \frac{1}{\mu_{a}^{2}} A_{i,\mu}^{'}
\end{align} 
Similarly, let $ \mu_{a} = (\frac{a_m}{a_{max}})^{\frac{1}{2}} $, then $ \mid \hat{A}_{i,\mu} \mid = \frac{a_{max}}{a_{m}} \mid A_{i,\mu}^{'} \mid = a_{max} \in \left[ -a_{max}, a_{max} \right] $.
}

\subsection{Iterative Time Adjustment}
\label{subs:iterative}
\begin{algorithm}
	\label{alg:adj}
	{
		\Repeat{$ \mathcal{V}.\textnormal{\textbf{empty}}() \land \mathcal{A}.\textnormal{\textbf{empty}}() $}
		{
			$ \mathcal{V}, \mathcal{A} = \textnormal{\textbf{findInfeasible}}() $\;
			\For{$ \mathbf{V}_{i}^{'} \ \textbf{in} \ \mathcal{V} $}{
				$ v_m \gets \max_{\mu \in \{x,y,z \}} \mid V_{i,\mu}^{'} \mid $\; 
				$ \mu_{v}^{'} \gets \min \{\alpha_v, \frac{v_m}{v_{max}} \} $\;
				\textbf{AdjustKnotSpansVel}($ \mathbf{V}_{i}^{'} $, $ \mu_v^{'} $)\;
			}
			\For{$ \mathbf{A}_{i}^{'} \ \textbf{in} \ \mathcal{A} $}{
				$ a_m \gets \max_{\mu \in \{x,y,z \}} \mid A_{i,\mu}^{'} \mid $\;
				$ \mu_{a}^{'} \gets \min \{\alpha_a, (\frac{a_m}{a_{max}})^{\frac{1}{2}} \} $\;
				\textbf{AdjustKnotSpansAcc}($ \mathbf{A}_{i}^{'} $, $ \mu_a^{'} $)\;
			}
		}
	}
	\caption{Iterative Time Adjustment}
\end{algorithm}
{
Based on the derivation in Sect.\ref{subs:adj}, Alg. \ref{alg:adj} is adopted to enforce dynamic feasibility.
It iteratively finds the infeasible velocity and acceleration control points $ \mathcal{V} $ and $ \mathcal{A} $ of the trajectory (Line 2) and adjust the corresponding knot spans (Lines 3-10). 
Because a knot span $ \Delta t_{m} $ influences a few control points and vice versa, bounding $ \mu_{v}^{'} $ and $ \mu_{a}^{'} $ with two constant $ \alpha_v $ and $ \alpha_a $ slightly larger than $ 1 $ (Line 5, 9) prevents any time span from being extended excessively.
}
\section{Implementation Details}
\label{sect:imp}

\subsection{Experiment Settings}
\label{subs:plat}
The motion planning method proposed in this paper is implemented in C++11 with a general non-linear optimization solver NLopt\footnote{\url{https://nlopt.readthedocs.io/en/latest/}}. 
{We set $ r=2, \tau = 0.5 $ for the path searching, $ \lambda_1 = 10.0, \lambda_2 = 0.8, \lambda_3 = 0.01 $ for the optimization and $ \alpha_a = \alpha_v = 1.1 $ for the time adjustment in all experiments.}
We present two sets of real-world experiments to validate our proposed planning method.
\begin{figure}[t]
	\begin{center}          
	\subfigure[\label{fig:drone1} Quadrotor platform used in the autonomous flight experiment.]
	{\includegraphics[width=0.45\columnwidth]{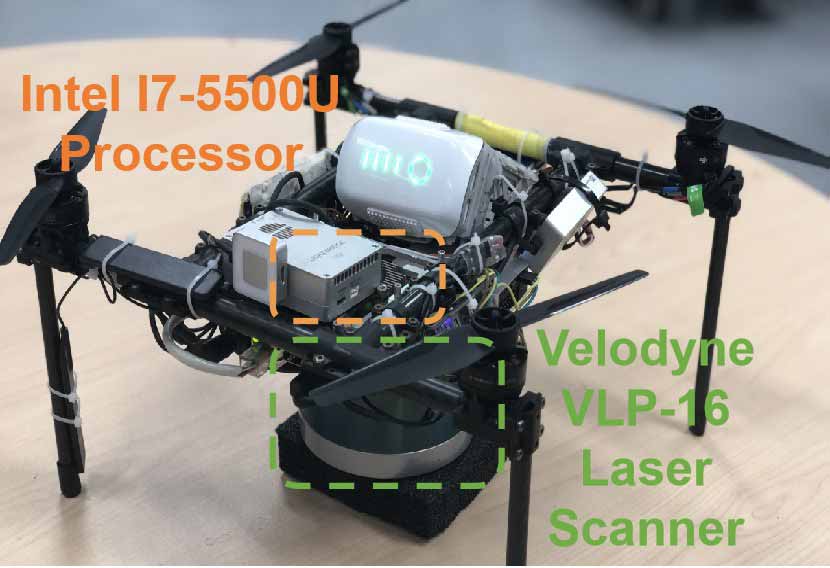}}       
	\subfigure[\label{fig:drone2} Quadrotor platform used in the aggressive replanning experiment.]
	{\includegraphics[width=0.45\columnwidth]{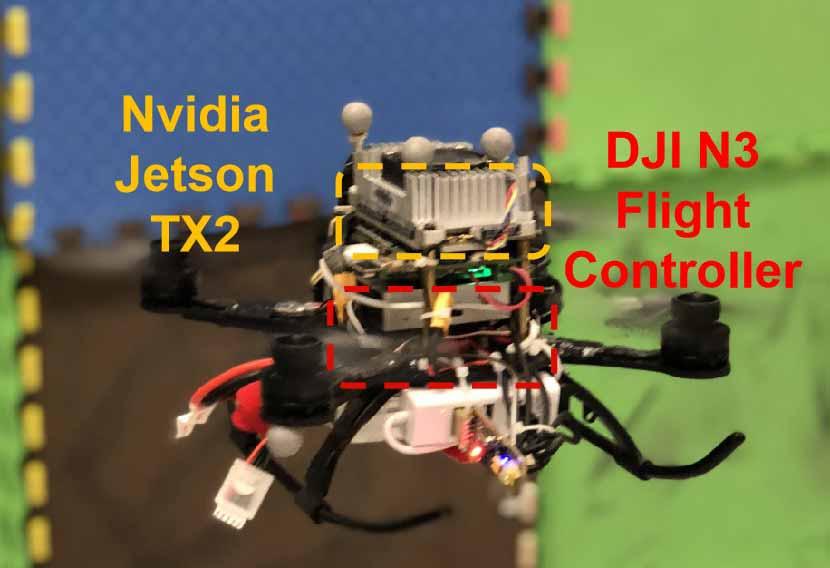}}     
	\end{center}
	\caption{\label{fig:drone} Quadrotor Platforms used in fully autonomous flight in (a), and aggressive kinodynamic replanning in (b). }
\vspace{-0.7cm}
\end{figure}  

Firstly, we conduct fast autonomous flight experiments in unknown cluttered environments (Sect.\ref{subs:onboard}). We use a self-developed quadrotor platform (Fig.~\ref{fig:drone1}) equipped with a Velodyne VLP-16 3-D LiDAR. LOAM\cite{zhang2014} is adopted to estimate the pose of the quadrotor and generate a dense point cloud map. To obtain high-rate state estimation for feedback control, we fused the laser-based estimation with IMU and sonar measurements by the extended Kalman filter (EKF). All modules including motion planning, state estimation, mapping and control run on a dual-core 3.00 GHz Intel i7-5500U processor, which has 8 GB RAM and 256 GB SSD.

{Then, in Sect.\ref{subs:mocap}, we focus on testing the fast-replanning capability of our proposed method in aggressive flight, for which we use a more light-weight and agile quadrotor platform (Fig.~\ref{fig:drone2})}. To eliminate uncertainties introduced by onboard sensings, accurate pose feedback is provided by the motion capture system \textit{OptiTrack}\footnote{\url{https://optitrack.com/}} and the map of the environment is pre-built. The motion planning and control modules run onboard on an Nvidia TX2 computer.
\vspace{-0.4cm}
\subsection{Re-planning Strategy}
\label{subs:plan_stra}

\subsubsection{Receding-horizon Local Planning}
\begin{figure}[t]
	\centering
	\includegraphics[width=0.6\columnwidth]{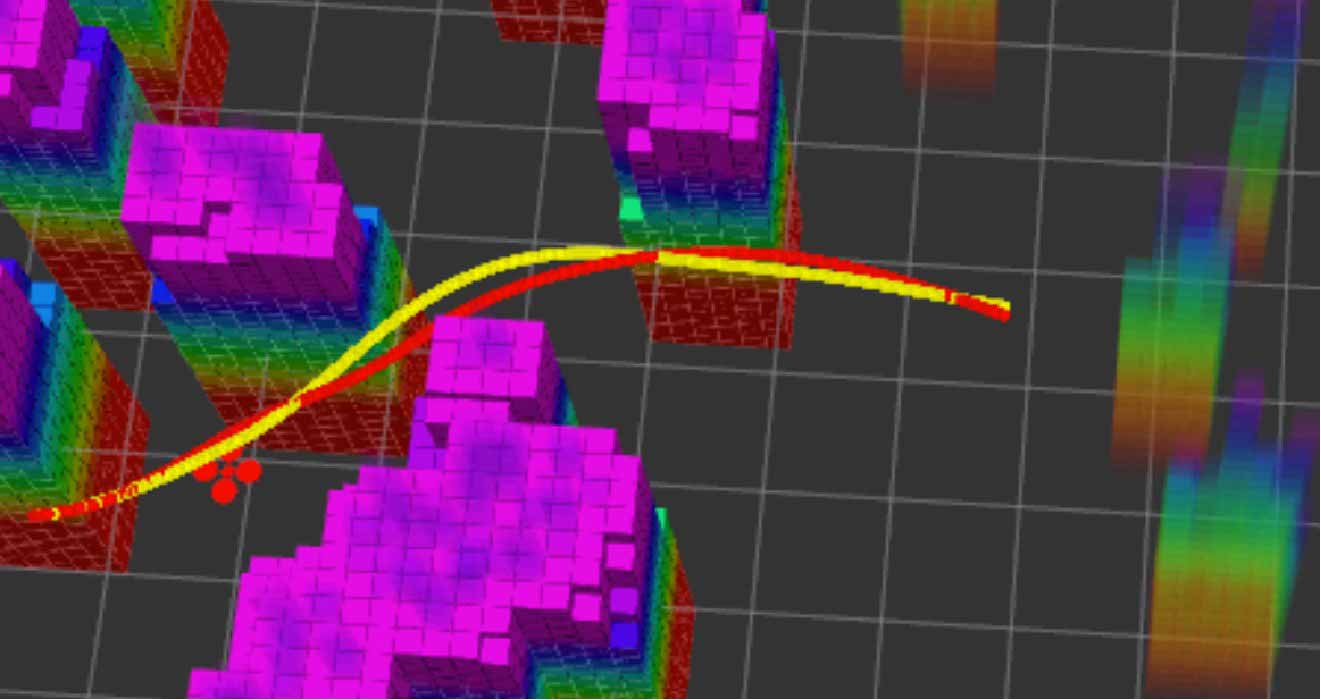} 
	\caption{The local planning strategy for a limited sensing range. The red curve and the yellow curve are the trajectories before and after the optimization. Opaque and transparent obstacles are known and unknown ones. \label{fig:replan_stra}} 
\vspace{-0.5cm}
\end{figure}

When the quadrotor flies in an unknown environment, it has to re-plan its trajectory frequently due to the limited sensing range. To improve efficiency, we adopt a receding-horizon planning scheme, in which trajectories are generated only within the known space (Fig.\ref{fig:replan_stra}). The path searching is terminated once a motion primitive ends outside this range and is followed by the optimization and time adjustment. Planning in the unknown space is often useless, thus such efforts can be saved.

\subsubsection{Re-planning Triggering Mechanism}
The re-planning is triggered in two situations. Firstly, it is triggered if the current trajectory collides with newly emergent obstacles\footnote{{The collision checking is conducted in the configuration space (C-space), where the quadrotor is modeled as a point and obstacles are inflated by a radius. The computation cost of such checking is negligible.} }, which ensures that a new safe trajectory is available as soon as any collision is detected. Secondly, the planner is called at fixed intervals of time. It updates the trajectory periodically using the most up-to-date environmental information. 
\vspace{-0.4cm}
\subsection{Euclidean Distance Field}
We maintain an EDF of the voxel grid map for our optimization, which is computed by an efficient $ O(n) $ algorithm~\cite{felzenszwalb2012distance}, where $ n = N^{3} $ is the number of updated voxel grids. 
To compensate for the discretized error of the EDF introduced by the voxel grid map and benefit the numeric optimization, trilinear interpolation is used to improve the accuracy of the distance and gradient information \cite{usenko2017real}. Global update of the EDF is very costly and can block the planning module that is crucial for fast autonomous flight. To address this issue, we only update the voxel grids within the sensing range using an incremental update strategy\cite{schouten2010incremental}.
\vspace{-0.5cm}
\section{Results}
\label{sect:res}

\subsection{Analysis and Comparisons}
\label{subs:compare}

\subsubsection{Comparison of Path Searching}
\label{ssubs:analy}
We compare our path searching with method \cite{liu2017iros}, both of which use the time-optimal control formulation to generate primitives. The comparison is done on a $40\times40\times5 m$ map randomly deployed with 100 obstacles {and the maximum velocity and acceleration limits are set as $3m/s$ and $ 2m/s^{2} $ respectively. Since the resolution of voxel grids is a critical factor for the performance of our proposed method, different resolutions are used for comprehensive evaluation (Tab.\ref{tab:1}, column 1, rows 3-5).} For a fair comparison, we use the open source implementation of \cite{liu2017iros}. 
Results are listed in {Tab.\ref{tab:1}}. 

{As is shown in statics, both methods generate kinodynamic feasible trajectories. Our method is faster with one order of magnitude and tends to generate a path with a shorter duration. However, the control cost for it is slightly higher. As the voxel girds get coarser, the efficiency of our method increases at the expense of higher control cost and lower success rate. This trend is expected because pruning primitives with coarser voxel grids results in lower searching complexity, whereas more feasible (and maybe superior) paths are lost.}

\begin{table}[t]
	\centering
	\caption{Comparison of Path Searching\label{tab:1}}
	\setlength{\tabcolsep}{2.5pt}
	{
	\begin{tabular}{|c|c|c|c|c|c|}
	\hline
	\multicolumn{2}{|l|}{}                                                                  & \begin{tabular}{c}Comp.  \\ Time(s)\end{tabular} & \begin{tabular}{c}Traj. \\ Time(s)\end{tabular} & \begin{tabular}{c}Ctrl. \\ Cost($m^2/s^3$)\end{tabular} & \begin{tabular}{c}Succ. \\ Rate(\%)\end{tabular} \\ \hline
	\multirow{3}{*}{method\cite{liu2017iros} }                                                 & Mean &                          0.0592                            &                          8.439                           &                              \textbf{11.42}                   & \multirow{3}{*}{100.0}                                     \\ \cline{2-5}
																																									 & Max  &                          0.8740                             &                           20.000                         &                              28.00                           &                                                       \\ \cline{2-5}
																																									 & Std  &                          0.1060                             &                         2.970                            &                              4.07                            &                                                       \\ \hline
	\multirow{3}{*}{\begin{tabular}{c}Proposed \\ (Res.= 0.04m)\end{tabular}}          & Mean &                          \textbf{0.0047}                    &                         \textbf{7.719}                   &                              \textbf{14.42}                  & \multirow{3}{*}{100.0}                                     \\ \cline{2-5}
																																									 & Max  &                          0.1837                             &                         10.500                           &                              24.50                           &                                                       \\ \cline{2-5}
																																									 & Std  &                          0.0113                             &                         0.909                            &                              3.60                            &                                                       \\ \hline
	\multirow{3}{*}{\begin{tabular}{c}Proposed \\ (Res.= 0.2m)\end{tabular}}           & Mean &                          \textbf{0.0018}                    &                         \textbf{7.696}                   &                              \textbf{15.08}                  & \multirow{3}{*}{100.0}                                     \\ \cline{2-5}
																																									 & Max  &                          0.0287                             &                         10.500                           &                              27.50                                &                                                       \\ \cline{2-5}
																																									 & Std  &                          0.0017                             &                         0.917                            &                              3.78                                &                                                       \\ \hline
	\multirow{3}{*}{\begin{tabular}{c}Proposed \\ (Res.= 1.0m)\end{tabular}}           & Mean &                          \textbf{0.0017}                    &                        \textbf{7.645}                    &                             \textbf{16.20}                             & \multirow{3}{*}{\textbf{77.8}}                                     \\ \cline{2-5}
																																									 & Max  &                          0.0059                            &                          15.500                             &                              56.00                              &                                                       \\ \cline{2-5}
																																									 & Std  &                          0.0007                             &                         1.295                          &                               6.21                               &                                                       \\ \hline
	\end{tabular}
	}
\end{table}
\subsubsection{Comparison of Optimization}
\label{ssubs:compare}
\begin{figure}[t]
	\begin{center}          
	\subfigure[\label{fig:back_cost} The convergent performance is compared through the cost profiles of the objective function. The cost of the proposed method decreases rapidly to zero within $ 3 ms $, while the other takes a significantly longer time. The cost for both methods are normalized to \text{[0,1]} for comparison. {Computation time is limited to $ 30 ms $ for both.}]
	{\includegraphics[width=0.9\columnwidth]{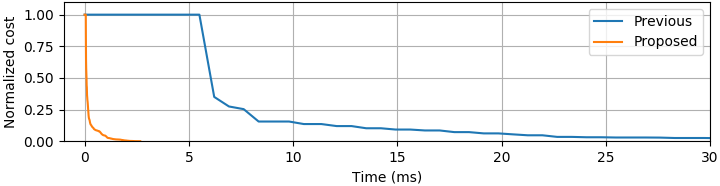}}       
	\subfigure[\label{fig:back_traj} {Trajectories generated by the proposed (red) and previous (blue) method. Note that computation time for the proposed method is only $1ms$, while that for the previous method is $10ms$.}]
	{\includegraphics[width=0.45\columnwidth]{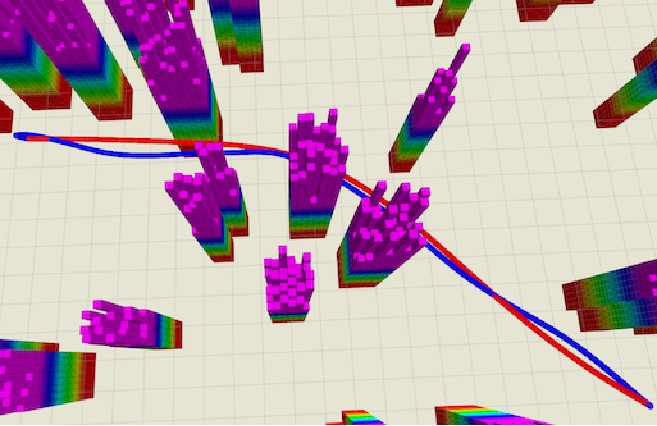}       
	\includegraphics[width=0.45\columnwidth]{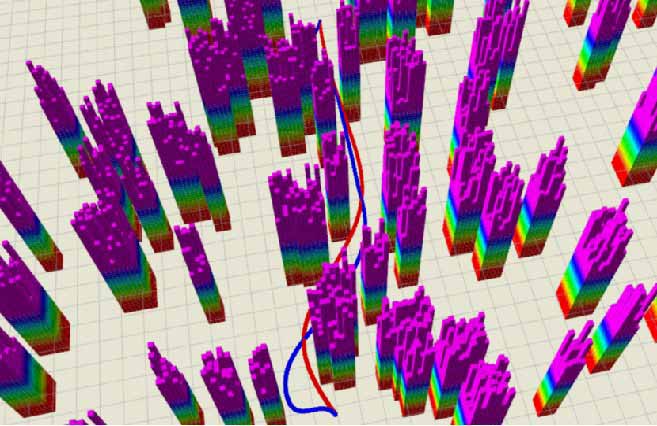}}     
	\end{center}
	\caption{\label{fig:back_compare} Comparing our proposed optimization method with gradient-based optimization method\cite{fei2017iros} in random cluttered environments.}
	\vspace{-0.2cm}
\end{figure}

For the back-end trajectory optimization, we conduct a comparison against our previous work\cite{fei2017iros}. Both of our previous method and the proposed method utilize the EDF for non-linear optimization. For fairness, we use the same path given by our path searching as the initial value. Firstly, we compare the costs of the objective function with respect to time for both methods (Fig.~\ref{fig:back_cost}). Obviously, the cost of the proposed method drops rapidly within the first few milliseconds, while the other one decreases much slower. Secondly, comparison of smoothness (integral of the squared jerk) is conducted as shown in Fig.\ref{fig:back_traj} and Tab.\ref{tab:back}. Even though less time is given for the proposed method, the resulting trajectories are smoother.

\begin{table}[t]
	\centering
	\caption{Comparison of Trajectory Optimization\label{tab:back}}
	\begin{tabular}{|c|c|c|c|c|} 
	\hline
	\multirow{2}{*}{} & \multicolumn{3}{c|}{{Integral of Jerk$^2$ ($m^{2}/s^{5})$}} & \multirow{2}{*}{{\begin{tabular}{c}Comp.  \\ Time(s)\end{tabular}}}  \\ 
	\cline{2-4}
										&   Mean   &   Max  &  Std  &  \\ 
	\hline
	Previous\cite{fei2017iros}          & 43.913 & 181.495 & 18.394 & {0.010} \\ 
	\hline
	Proposed          & \textbf{35.932} & \textbf{131.913} & \textbf{13.118} & {0.001} \\
	\hline
	\end{tabular}
	\vspace{-0.5cm}
\end{table}

\subsection{Onboard Autonomous Flight}
\label{subs:onboard}
\begin{figure}[t]
	\begin{center}          
	\subfigure[\label{fig:env} The environment setup of the autonomous flight experiment.]
	{\includegraphics[width=0.447\columnwidth]{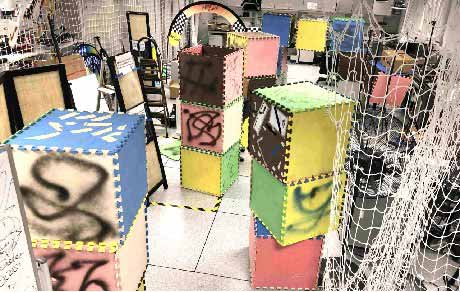}}       
	\subfigure[\label{fig:bag1} Flight 1, bypassing a vertical wall.]
	{\includegraphics[width=0.447\columnwidth]{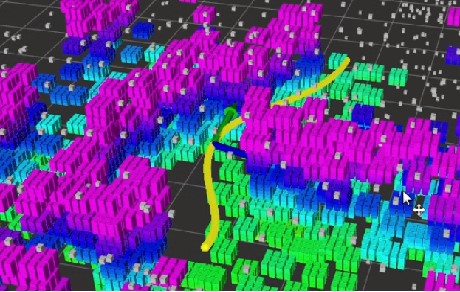}}       
	\vspace{-0.3cm}
	\subfigure[\label{fig:bag2} Flight 2, avoiding vertical obstacles.]
	{\includegraphics[width=0.447\columnwidth]{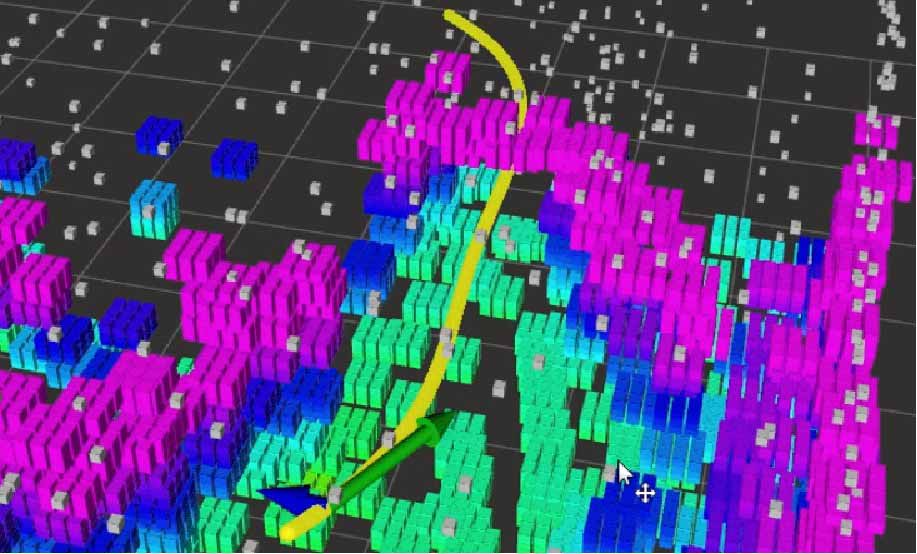}}       
	\subfigure[\label{fig:bag3} Flight 3, flying through a narrow hole.]
	{\includegraphics[width=0.447\columnwidth]{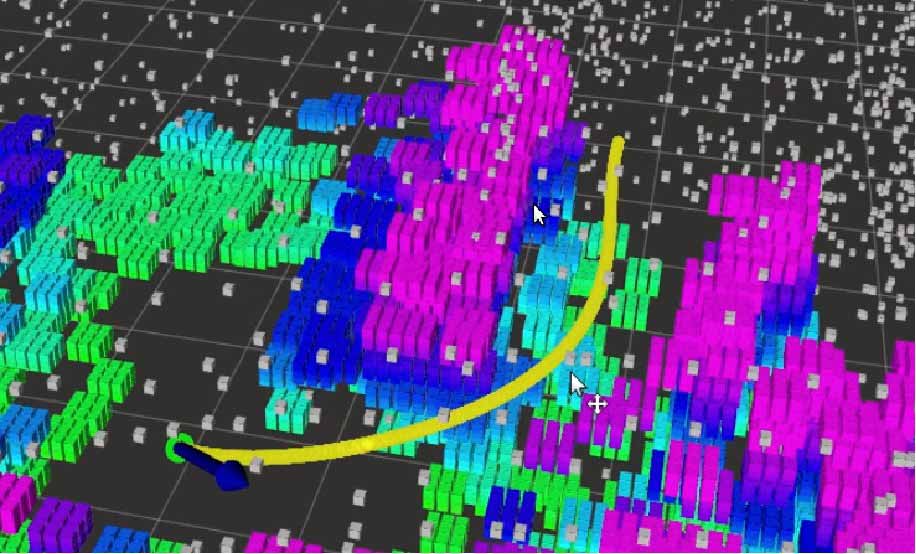}}     
	\end{center}
	\caption{\label{fig:indoor_bag} Fully autonomous flight in an unknown cluttered environment. Only the colored map is known by the motion planning module. In this confined environment, the maximum and average speed of {flight 1-3 reach up to $ 1.7 m/s $ and $ 1.3 m/s^2 $} respectively.}
	\vspace{-0.3cm}
\end{figure}

We conducted fully autonomous fast flight experiments in a challenging unknown environment (Fig.~\ref{fig:env}). To further challenge our method, we prune the global map using a sphere with a radius of $ 5 \ m $ that centered on the quadrotor and only use the map within this sphere for trajectory generation (Fig.\ref{fig:bag1}-\ref{fig:bag3}) which is much smaller than our real perception range. The unstructured environment, limited perception range as well as the high flying speed pose a challenge to the motion planning module, as it should re-generate trajectory continually and rapidly upon sudden appearing of new threats. We refer the readers to the video attachment for more detailed information.

\subsection{Aggressive Flight}
\label{subs:mocap}
\begin{figure}[t]
	\begin{center}          
	\subfigure
	{\includegraphics[width=0.447\columnwidth]{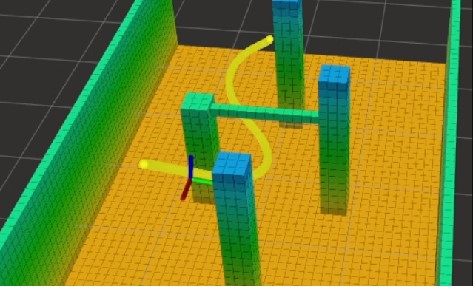}}       
	\subfigure
	{\includegraphics[width=0.447\columnwidth]{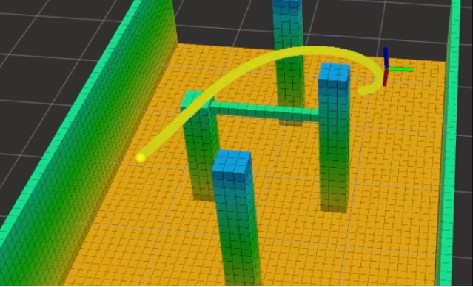}}     
	\vspace{-0.5cm}
\end{center}
\caption{\label{fig:agg_bag} Aggressive flight test. The goals are changed arbitrarily and new trajectories are replanned during the aggressive flight.  }
	\vspace{-0.4cm}
\end{figure}

The aggressive flight experiment is done in the environment depicted in Fig.~\ref{fig:agg_bag}. In the experiment, the goals of the quadrotor are changed constantly and arbitrarily by a human. As soon as a new goal is set, a new trajectory is re-planned and executed immediately. {The maximum velocity and acceleration are set as $ 2.5m/s $ and $ 1.5m/s^2 $ respectively. This task is challenging in several aspects. Since the flight is aggressive and the changes in the goals are abrupt, the motion planning module should generate new trajectories in considerably short time to quickly react to the changes, so that the motion of the quadrotor is continuous and smooth. Also, as the environment is confined and cluttered, it is difficult to generate smooth, safe and dynamically feasible trajectories in a very short time.} 
This experiment validates that our method can generate aggressive motion under the premise of feasibility. It also shows that our method can quickly generate a new trajectory in complex environments even if the goal is changed suddenly during the aggressive flight. More details are also included in the video.

\vspace{-0.05cm}
\section{Conclusion}
\label{sect:con}
In this paper, we propose a novel online motion planning method for quadrotor autonomous navigation. We decouple the online fast motion planning problem as a front-end kinodynamic path searching and a back-end nonlinear trajectory optimization. We adopt a kinodynamic path searching to find a safe, kinodynamic feasible and minimum-time initial path, which is further improved in smoothness and clearance by a gradient-based optimization. By utilizing the convex hull property of B-spline, we significantly improve the efficiency and convergent rate of the optimization compared to previous gradient-based planning methods. Finally, by representing the trajectory as a non-uniform B-spline, we adjust the time allocation according to a given expected flight aggressiveness. We validate our proposed method in various complex environments and the simulation. The competence of the method is also validated in challenging real-world tasks.

In the future, we plan to challenge our quadrotor system in extreme situations such as large-scale or dynamic environments. Furthermore, we will extend our trajectory optimization method to swarm problems.


\newlength{\bibitemsep}\setlength{\bibitemsep}{0.11\baselineskip}
\newlength{\bibparskip}\setlength{\bibparskip}{0pt}
\let\oldthebibliography\thebibliography
\renewcommand\thebibliography[1]{%
  \oldthebibliography{#1}%
  \setlength{\parskip}{\bibitemsep}%
  \setlength{\itemsep}{\bibparskip}%
}
\bibliography{zby} 
\end{document}